\documentclass[11pt]{article}

\usepackage[final]{acl}

\usepackage{times}
\usepackage{latexsym}
\usepackage{amssymb}

\usepackage[table]{xcolor}
\usepackage{booktabs}
\usepackage{array}

\usepackage[T1]{fontenc}

\usepackage[utf8]{inputenc}

\usepackage{microtype}

\usepackage{inconsolata}

\usepackage{graphicx}
\usepackage{booktabs}
\usepackage{amsmath}
\usepackage{multirow}
\usepackage[table]{xcolor}
\usepackage{tikz}
\usepackage[table]{xcolor}
\usepackage{comment}

\definecolor{cIncorrect}{HTML}{EB6834}
\definecolor{cSecond}{HTML}{2A78D6}
\definecolor{cSwap}{HTML}{1BAF7A}
\definecolor{cMism}{HTML}{8A8A85}

\newcommand{\mathdltup}[1]{\textcolor{green!60!black}{\ensuremath{\uparrow}\,#1}}
\newcommand{\mathdltdn}[1]{\textcolor{red!70!black}{\ensuremath{\downarrow}\,#1}}
\newcommand{\mathdltnz}[1]{\textcolor{gray}{#1}}
\newcommand{\ci}[1]{{\scriptsize$\pm$#1}}

\newcommand{\rparagraph}[1]{\vspace{1.4mm}\noindent\textbf{#1}}

\newcommand{\sparagraph}[1]{\vspace{0.0mm}\noindent\textbf{#1}}

\newcommand*\circled[1]{\tikz[baseline=(char.base)]{
            \node[shape=circle,draw,inner sep=.6pt] (char) {#1};}}

\title{On-Policy Self-Distillation Does Not Require \\ Verified Privileged Information}

\author{
\textbf{Gorjan Radevski}$^{2}$\thanks{Correspondence to: gorjan.radevski@gmail.com or kiril.gashteovski@neclab.eu},
\textbf{Kiril Gashteovski}$^{1}$,
\textbf{Sascha Saralajew}$^{1}$,\\
\textbf{Gürkan Solmaz}$^{1}$,
\textbf{Ammar Shaker}$^{1}$,
\textbf{Carolin Lawrence}$^1$,
\textbf{Goran Glavaš}$^3$ 
 \\
$^1$NEC Laboratories Europe, Germany
$^2$Independent \\
$^3$Center for Artificial Intelligence and Data Science, University of Würzburg, Germany
}

\begin{document}
\maketitle
\begin{abstract}
On-policy self-distillation (OPSD) trains a large language model on its own rollouts, with targets produced by the same model (as a teacher), while conditioned on \textit{privileged information}. The privileged information is typically \textit{verified} reference solutions or reasoning trajectories.
In this focused contribution, we investigate whether the privileged trajectory must be correct (i.e., conclude with a correct answer). Our experiments on text-only and vision-language mathematical reasoning render the correctness of privileged trajectories from a (larger) expert mostly inconsequential for OPSD, suggesting that an OPSD student benefits from the expert's more advanced reasoning even when incorrect. 
The same, however, does not hold for supervised fine-tuning: SFT-ing on incorrect expert trajectories yields significantly weaker students than training on correct ones.  
To explain this asymmetry, we probe the teacher's training signal and find that correct and incorrect contexts produce nearly the same per-token update directions, diverging mainly on the final answer tokens, which carry little of the loss.
\end{abstract}
\section{Introduction}
On-Policy Self-Distillation (OPSD) trains a model on its own rollouts using token-level supervision from itself, with the teacher instance conditioned on privileged information such as a reference solution \cite{zhao2026self, shenfeld2026self} or verified trajectories \cite{nguyen2026avsd}. 
The idea is that the model ``internalizes'' the privileged context and teaches its weaker, unconditioned self. 
Most recent works show the success of OPSD in text-only \cite{zhao2026self, shenfeld2026self, mendes2026making}, multimodal \cite{tian2026vicur, liang2026visual, xiang2026teaching}, and omni-modal settings \cite{zhao2026opod}, but they all rely on \textit{verified} solutions \cite{zhao2026self, mendes2026making} or (parts of)
verified trajectories \cite{nguyen2026avsd} as privileged context.\footnote{See Appendix~\ref{app:related} for positioning w.r.t. related work.}

In this focused work, we ask \textit{whether the privileged trajectory must conclude with a correct answer} for OPSD to work. 
We first sample reasoning trajectories for both text-only and vision-language math reasoning from a stronger expert. We then condition the OPSD teacher on correct \textit{or} incorrect expert trajectories and compare against supervised fine-tuning (SFT) on the same trajectories. We find that OPSD yields similar students under either condition, while SFT does not: SFT-ing on incorrect trajectories results in tangibly weaker students. 


We find that a correct and an incorrect trajectory induce very similar updates 
on the student's tokens, much like updates from two independent correct trajectories (and far more than a trajectory of another problem). Their directions diverge more on the last few tokens of the student's rollout, 
which make up a small fraction of generated tokens. 
This explains the difference between the objectives: SFT imitates the trajectory itself, whereas OPSD uses it to set targets on the student's own tokens, where correct and incorrect trajectories largely agree.

\rparagraph{Contributions.}
\circled{1} We show that OPSD yields similar students under both correct and incorrect expert trajectories as privileged context, while SFT on the same trajectories does not (\S\ref{sec:main}).
\circled{2} We locate the gains: they concentrate on problems the expert solves and the base model cannot; here, OPSD improves twice as much as SFT, without degrading on problems the base model can solve (\S\ref{sec:partition}).
\circled{3} We probe the teacher's signal and find that correct and incorrect contexts produce nearly the same per-token update directions, diverging mainly on the late and answer tokens that carry little of the loss; an unrelated expert trajectory, as a control, produces a qualitatively different signal (\S\ref{sec:probe}).

\section{Setup}\label{sec:method}
We follow the standard OPSD recipe \cite[\textit{inter alia}]{zhao2026self, shenfeld2026self, kaur2026rethinking}: a single model with parameters $\theta$ acts as two policies, differing only in the input they get:
\begin{equation*}
\underbrace{\pi_\theta(\cdot \mid x, c)}_{\text{teacher}}
\qquad\text{and}\qquad
\underbrace{\pi_\theta(\cdot \mid x)}_{\text{student}},
\end{equation*}
where $x$ is the input (e.g., a math question or an image-question pair) and $c$ is the \emph{privileged context} available only to the teacher in training. At inference, the model sees only $x$.

\rparagraph{Objective.}
Given a student rollout $y \sim \pi_\theta(\cdot \mid x)$, we minimize the per-token forward KL from the teacher to the student over the response tokens:
\vspace{-0.7em}

{
\small
\begin{equation*}
\mathcal{L}_{\mathrm{SD}}(\theta)
= \mathbb{E}_{(x,c)}\,\mathbb{E}_{y \sim \pi_\theta(\cdot \mid x)}
\left[\frac{1}{|y|}\sum_{t=1}^{|y|} \mathrm{KL}\!\left(q_t \,\Vert\, p_t\right)\right],
\end{equation*}
}

\noindent with $q_t \hspace{-0.2em} = \hspace{-0.2em} \pi_{\bar\theta}(\cdot \hspace{-0.2em} \mid x, c, y_{<t})$ and $p_t \hspace{-0.2em} = \hspace{-0.2em} \pi_{\theta}(\cdot \hspace{-0.2em} \mid x, y_{<t})$ as respective next-token distributions of the teacher and student and $\bar\theta$ a stop-gradient copy of $\theta$. As the teacher and student share all weights, the only difference between $q_t$ and $p_t$ is the privileged context $c$: this is the supervision signal in OPSD. 

\rparagraph{Privileged context.} Contexts $c$ are reasoning traces, which we vary along two dimensions: (1) \emph{Source:} $c$ is sampled from a large expert $c \sim \pi_{\mathrm{E}}(\cdot \mid x)$ \textit{or} from the model itself $c \sim \pi_{\theta_0}(\cdot \mid x)$;  \emph{Correctness:} $c$ is \emph{correct} ($c^{\checkmark}$) if its final answer matches the reference answer and \emph{incorrect} ($c^{\times}$) otherwise. Crucially, existing work \cite{zhao2026self, shenfeld2026self} uses only verified contexts $c^{\checkmark}$.

\rparagraph{SFT.} 
To decouple the effect of $c$ as \emph{privileged context} from the impact of \emph{training data}, we train on the exact same trajectories via supervised fine-tuning (SFT): $\mathcal{L}_{\mathrm{SFT}} = -\log \pi_\theta(c \hspace{-0.2em}\mid\hspace{-0.2em} x)$. OPSD and SFT thus consume identical trajectories, with the only difference being that self-distillation uses them to condition the teacher, whereas SFT pushes the model to imitate them directly.

\rparagraph{Controls.} To ensure that the differences between setups are not due to \textit{which problems} a model sees, we train on problems for which the expert produced both a correct and an incorrect trajectory, so that $c^{\checkmark}$ and $c^{\times}$ represent (candidate) solutions for the same problems. In each setup, we train on the same $N$ problems for a single epoch with identical hyperparameters (Appendix~\ref{app:implementation}).
\section{Experiments \& Findings}

\sparagraph{Models.} 
We experiment with Qwen3-4B for text-only OPSD\footnote{A hybrid thinking model, like in \citet{zhao2026self}.} and with Qwen3-VL-4B-Instruct for vision-language OPSD, keeping the vision encoder frozen.
In both cases, we only LoRA fine-tune \cite{hu2022lora} the underlying LLM. The expert from which we sample privileged contexts is Qwen3.6-35B-A3B. In some experiments, we train with trajectories sampled from the base model itself ($\pi_{\theta_0}$), to which we refer as \textit{self-context}.

\rparagraph{Data and trajectory sampling.} We use DeepScaler \cite{deepscaler2025} for text-only training and MathInstruct \cite{wang2025mathcodervl, wang2024mathcoder} for vision-language reasoning. For each training problem, we sample $k_t=3$ expert trajectories and $k_s=1$ base-model trajectory; An LLM judge (Qwen3.6-35B-A3B) labels each trajectory as correct or incorrect by comparing its final answer against the reference (see Appendix~\ref{app:llm-judge} for details). 

\rparagraph{Evaluation.} We report accuracy on AIME24, AIME25, and HMMT2025 for text-only problems (avg@12: following \citet{zhao2026self}); and on MathVerse \cite{zhang2024mathverse}, MathVision \cite{wang2024measuring}, MathVista \cite{lu2024mathvista}, WeMath \cite{qiao2024we}, and DynaMath \cite{zou2024dynamathdynamicvisualbenchmark} for vision-language problems (avg@3).

\subsection{Main results}\label{sec:main}
Table~\ref{tab:math-reasoning} shows that OPSD with expert context gives the largest gains on both text-only and vision-language problems. Crucially, OPSD seems largely insensitive to the context's correctness: incorrect trajectories match correct ones for text-only problems (+1.95 for both) and trail them only slightly on vision-language problems (+2.98 vs.\ +3.69). 
In contrast, SFT gains on correct vision-language traces are more modest (+2.06), and this halves for incorrect trajectories (+0.93). For text-only problems, SFT on expert trajectories is unreliable: imitating correct trajectories degrades performance (-5.65), suggesting a confound: the behavior is not explained by correctness alone. 
Finally, the source of the context matters more than its correctness: with the base model's own trajectories, both OPSD and SFT gain around one point, well below self-distillation with expert context. This proves that OPSD gains indeed stem from expert trajectories and are not an effect of training problem selection.

\begin{table*}[t]
\centering
\def\arraystretch{0.9}
\resizebox{1.0\textwidth}{!}{%
\begin{tabular}{lcccccccccccc}
\toprule
Method &
Source &
Correct &
AIME24 &
AIME25 &
HMMT25 &
\cellcolor{gray!15} Avg. Text &
MathVerse &
MathVision &
MathVista &
WeMath &
DynaMath &
\cellcolor{gray!15} Avg. Vis.-Lang. \\
\midrule

Base &
-- &
-- &
70.83 &
68.61 &
43.88 &
\cellcolor{gray!15} 61.10 &
61.17 &
51.48 &
75.40 &
57.05 &
66.81 &
\cellcolor{gray!15} 62.38 \\

Expert &
-- &
-- &
80.00 &
62.78 &
61.11 &
\cellcolor{gray!15} 67.96 &
76.80 &
70.76 &
86.30 &
81.43 &
76.71 &
\cellcolor{gray!15} 78.40 \\

\midrule

\rowcolor{cSecond!8}
SFT &
Expert &
$\checkmark$ &
\mathdltdn{7.22} &
\mathdltdn{7.23} &
\mathdltdn{2.50} &
\cellcolor{gray!15} \mathdltdn{5.65} &
\mathdltup{3.02} &
\mathdltup{1.58} &
\mathdltup{1.40} &
\mathdltup{4.47} &
\mathdltdn{0.18} &
\cellcolor{gray!15} \mathdltup{2.06} \\

\rowcolor{cIncorrect!8}
SFT &
Expert &
$\times$ &
\mathdltup{1.94} &
\mathdltdn{3.61} &
\mathdltup{3.34} &
\cellcolor{gray!15} \mathdltup{0.56} &
\mathdltup{3.88} &
\mathdltup{2.43} &
\mathdltup{0.90} &
\mathdltdn{1.91} &
\mathdltdn{0.64} &
\cellcolor{gray!15} \mathdltup{0.93} \\

\rowcolor{cSecond!8}
Self-Distillation &
Expert &
$\checkmark$ &
\mathdltup{4.44} &
\mathdltdn{0.56} &
\mathdltup{1.95} &
\cellcolor{gray!15} \mathdltup{1.95} &
\mathdltup{5.28} &
\mathdltup{2.73} &
\mathdltup{1.60} &
\mathdltup{7.24} &
\mathdltup{1.57} &
\cellcolor{gray!15} \mathdltup{3.69} \\

\rowcolor{cIncorrect!8}
Self-Distillation &
Expert &
$\times$ &
\mathdltup{5.00} &
\mathdltup{1.39} &
\mathdltdn{0.55} &
\cellcolor{gray!15} \mathdltup{1.95} &
\mathdltup{5.58} &
\mathdltup{2.24} &
\mathdltup{1.70} &
\mathdltup{4.47} &
\mathdltup{0.91} &
\cellcolor{gray!15} \mathdltup{2.98} \\

\midrule

\rowcolor{cSecond!8}
SFT &
Self &
$\checkmark$ &
\mathdltup{3.61} &
\mathdltdn{4.17} &
\mathdltup{3.34} &
\cellcolor{gray!15} \mathdltup{0.93} &
\mathdltup{4.06} &
\mathdltup{3.19} &
\mathdltup{1.10} &
\mathdltdn{2.19} &
\mathdltdn{0.24} &
\cellcolor{gray!15} \mathdltup{1.19} \\

\rowcolor{cIncorrect!8}
SFT &
Self &
$\times$ &
\mathdltup{3.61} &
\mathdltdn{2.78} &
\mathdltup{0.57} &
\cellcolor{gray!15} \mathdltup{0.47} &
\mathdltup{4.59} &
\mathdltup{3.16} &
\mathdltup{0.50} &
\mathdltdn{1.05} &
\mathdltdn{0.98} &
\cellcolor{gray!15} \mathdltup{1.25} \\

\rowcolor{cSecond!8}
Self-Distillation &
Self &
$\checkmark$ &
\mathdltup{2.22} &
\mathdltdn{1.67} &
\mathdltup{1.95} &
\cellcolor{gray!15} \mathdltup{0.83} &
\mathdltup{2.54} &
\mathdltup{0.23} &
\mathdltup{1.00} &
\mathdltup{1.33} &
\mathdltup{0.02} &
\cellcolor{gray!15} \mathdltup{1.02} \\

\bottomrule
\end{tabular}%
}
\caption{Qwen3-4B student for language only as per \citet{zhao2026self} and a Qwen3-VL-4B-Instruct student for vision-language; Qwen3.6-35B-A3B expert. Base and Expert report absolute performance; all other entries report deltas---\textcolor{green!60!black}{positive ($\uparrow$)} or \textcolor{red!70!black}{negative ($\downarrow$)} with respect to Base performance. \textit{Source}: the model that provided the privileged context; \textit{Correct} indicates if the privileged context is correct. All results are averages over 3 runs. 
}
\label{tab:math-reasoning}
\vspace{-0.5em}
\end{table*}

\subsection{Where do the gains come from?}\label{sec:partition}
\vspace{-0.3em}

Accuracy alone does not profile the problems on which OPSD improves. To gain this insight, we partition the vision-language test problems by whether the base model (before any OPSD training!) and the expert correctly solve them: (i) problems both correctly solve ($B~\checkmark E~\checkmark$); (ii) problems only the expert solves ($B~\times E~\checkmark$); (iii) problems neither solves ($B~\times E~\times$); and (iv) problems only the base model solves ($B~\checkmark E~\times$).
We create a fixed assignment of problems to bins based on one generated sample from the Base and one from the Expert. At inference, we stochastically sample solutions for all models, including the Base itself: because of this (i.e., re-sampling), Base \textit{can} score below 100\% on $B~\checkmark$ bins and above 0\% on $B~\times$ bins.
Comparing trained models against 100\% and 0\% that define the bins would thus mistake sampling noise for an improvement or a degradation. Instead, we compare against the mean of three independent resamples of the Base model. Table~\ref{tab:vlm-partition} reports the model performance for all four bins.

For problems that both models solve ($B~\checkmark E~\checkmark$), neither OPSD nor SFT change performance ($\leq 0.4$ points): both objectives seem to maintain the model's reasoning abilities. 
The gains come from problems the expert solves, but the base model does not ($B~\times E~\checkmark$): OPSD improves accuracy by 11.8\% with correct and 11.1\% with incorrect context; SFT gains less than half as much on correct traces. 
We further note that correctness affects the two objectives differently: for SFT, training on incorrect trajectories cuts the gains (from 5.3\% to 2.1\%); for OPSD, the 0.7\% gap is within noise. 
When an incorrect trajectory is used as an imitation target in SFT, the errors are learned; when it only conditions the teacher in OPSD, they are not. 
On problems neither model solves ($B~\times E~\times$), both objectives improve by 3--5\% regardless of correctness: we hypothesize that this is a generic effect of ``more reasoning training'' rather than a successful transfer of the expert's reasoning abilities. 

\begin{table}[t]
\centering
\small
\setlength{\tabcolsep}{4pt}
\renewcommand{\arraystretch}{0.95}
\resizebox{0.9\columnwidth}{!}{%
\begin{tabular}{lccccc}
\toprule
\multirow{2}{*}{Bin ($n$)} &
\multirow{2}{*}{Base} &
\multicolumn{2}{c}{SFT} &
\multicolumn{2}{c}{Self-Distillation} \\
\cmidrule(lr){3-4}
\cmidrule(lr){5-6}
& & $\checkmark$ & $\times$ & $\checkmark$ & $\times$ \\
\midrule

$B~\checkmark E~\checkmark$ (8,761) &
91.4 &
\mathdltnz{$-$0.2} &
\mathdltnz{+0.2} &
\mathdltnz{\phantom{+}0.0} &
\mathdltnz{$-$0.4} \\

\rowcolor{cSecond!12}
$B~\times E~\checkmark$ (3,044) &
26.6 &
\mathdltup{+5.3} &
\mathdltup{+2.1} &
\mathdltup{\textbf{+11.8}} &
\mathdltup{\textbf{+11.1}} \\

\rowcolor{cIncorrect!12}
$B~\times E~\times$ (2,466) &
10.0 &
\mathdltup{+4.0} &
\mathdltup{+2.9} &
\mathdltup{+5.1} &
\mathdltup{+5.2} \\

\rowcolor{cMism!12}
$B~\checkmark E~\times$ (906) &
69.9 &
\mathdltnz{$-$3.7} &
\mathdltnz{+2.0} &
\mathdltnz{$-$1.5} &
\mathdltnz{$-$3.1} \\

\bottomrule
\end{tabular}%
}
\caption{Performance on vision-language problems partitioned by whether the base ($B$) and the expert ($E$) can correctly solve them. 
\emph{Base}: the mean of three independent resamples of the base model; other columns report percentage-point changes relative to it. Shaded: bins in which post-trained models show relative gains. \textcolor{gray!80}{Gray}: change within two standard errors from Base.}
\label{tab:vlm-partition}
\vspace{-1em}
\end{table}

\begin{table}[t]
\centering
\footnotesize
\setlength{\tabcolsep}{5pt}
\renewcommand{\arraystretch}{0.95}
\resizebox{0.9\columnwidth}{!}{%
\begin{tabular}{lcccc}
\toprule
& & \multicolumn{3}{c}{Cosine\ with $\checkmark$} \\
\cmidrule(lr){3-5}
Context & Signal & All & Last 10\% & Ans. \\
\midrule

\multicolumn{5}{l}{\textbf{\textit{Vision-Language}}} \\
Correct ($\checkmark$)
  & 0.15 & -- & -- & -- \\
\rowcolor{cSecond!12}
Second correct
  & 0.15 & 0.77 & 0.74\ci{.02} & 0.73\ci{.03} \\
\rowcolor{cIncorrect!15}
Incorrect ($\times$)
  & 0.16 & 0.72 & 0.67\ci{.02} & 0.71\ci{.05} \\
\rowcolor{cSwap!12}
Answer-swapped
  & 0.15 & 0.98 & 0.90\ci{.01} & 0.88\ci{.07} \\
\rowcolor{cMism!15}
Mismatched
  & 0.04 & 0.34 & 0.36\ci{.02} & 0.43\ci{.10} \\

\midrule
\multicolumn{5}{l}{\textbf{\textit{Text}}} \\
Correct ($\checkmark$)
  & 0.19 & -- & -- & -- \\
\rowcolor{cSecond!12}
Second correct
  & 0.19 & 0.89 & 0.94\ci{.02} & 0.91\ci{.05} \\
\rowcolor{cIncorrect!15}
Incorrect ($\times$)
  & 0.19 & 0.86 & 0.88\ci{.03} & 0.88\ci{.05} \\
\rowcolor{cSwap!12}
Answer-swapped
  & 0.19 & 0.99 & 0.96\ci{.02} & 0.93\ci{.04} \\
\rowcolor{cMism!15}
Mismatched
  & 0.09 & 0.63 & 0.75\ci{.05} & 0.75\ci{.07} \\
\bottomrule
\end{tabular}
}
\caption{Teacher signals on the student's own rollouts
(500 training problems with at least two correct and one
incorrect expert trajectory).
\emph{Signal}: mean per-token $\mathrm{KL}(q_t\Vert p_t)$.
\emph{Cos.}: agreement between the update direction
$(q_t-p_t)$ of the teacher with a correct context and
with each other context, weighted by update size,
over all $y$ tokens (\emph{All}); the last 10\% of
$y$ tokens (\emph{Last 10\%}; rollouts ending within
the length limit); and $y$'s final-answer span
(\emph{Ans.}; rollouts with an answer marker).}
\label{tab:probe}
\vspace{-1em}
\end{table}

\subsection{Probing the Teacher's Signal}\label{sec:probe}
To understand why correct and incorrect contexts yield similar OPSD students, we measure the training signal directly. We sample 500 training problems for which the expert produced at least two correct and one incorrect trajectory, and take one rollout $y$ of the Base student for each. Then, the frozen Base model scores each token of $y$ under the student prompt ($p_t$), and under five different teacher contexts:
(i) a correct expert trajectory ($q^{\checkmark}_t$);
(ii) a second, independent correct trajectory, as a reference for two equally good contexts; (iii) an incorrect trajectory ($q^{\times}_t$) from the same problem;
(iv) an \emph{answer-swapped} context: the correct trajectory with its final answer replaced with an incorrect one; and (v) a correct trajectory of a different problem (\emph{mismatched}), as an irrelevant context reference.\footnote{Prompts, KL temperature and max response length match training. \textit{Concluding statement:} the text from the first answer marker (\texttt{\textbackslash boxed}, ``Final Answer'', ``Answer:'') in the last 500 characters; problems whose trajectories lack one are excluded.}

We report three quantities (each with a 95\% bootstrap interval over problems):
(i) \emph{Signal}: the mean per-token loss $\mathrm{KL}(q_t \Vert p_t)$.
(ii) \emph{Agreement}: at each token, training moves the student's logits along $q_t - p_t$; we report the cosine similarity between this update under the correct context ($q^{\checkmark}_t - p_t$; a trajectory ending in the right answer) and under each of the other four contexts: second correct, incorrect ($q^{\times}_t - p_t$),
answer-swapped, and mismatched. We average over tokens.
We report results in Table~\ref{tab:probe}.

\rparagraph{An incorrect context is nearly as aligned as a second correct one.}
Correct and incorrect contexts pull the student equally hard (\emph{Signal}: 0.15 vs.\ 0.16 for vision-language, 0.19 for both for text-only) and in largely the same direction. The agreement between two correct trajectories is not perfect because correct trajectories may follow different solution paths (cosine 0.77 for vision-language, 0.89 for text).
An incorrect trajectory is only slightly (and insignificantly) less aligned with the correct one (\emph{All}: vision-language cosine is 0.72, -0.05 less; text-only is 0.86, -0.03 less), and far more than an unrelated trajectory (0.34 and 0.63, resp.).

\rparagraph{The remaining difference is systematic, but confined to a few tokens.}
The gap between an incorrect and a second correct context widens toward the end of the rollout, where the incorrect trajectory's errors and final answer are.
For vision-language problems, the incorrect context's cosine drops from 0.72 (all tokens) to 0.67 (last 10\%), while a second correct context drops less, from 0.77 to 0.74. The same pattern holds for text-only problems (0.86$\to$0.88 vs.\ 0.89$\to$0.94). Tokens of the final answer drive this late divergence: swapping only the answer barely changes the update over all tokens (cosine 0.98 and 0.99) but lowers it in the last 10\% (0.90 and 0.96) and on the answer span (0.88 and 0.93). These few tokens carry little of the loss: the answer span accounts for 2.3\% and 0.7\% of the tokens in vision-language and text-only problems, yet 69\% of the rollouts reach the 1,024-token limit before stating an answer \citep{zhao2026self}. This, we believe, explains the asymmetry from Table~\ref{tab:vlm-partition}: SFT imitates every token in the trajectory, including its erroneous final steps, whereas self-distillation uses the trajectory only to shape targets on the student's own tokens, most of which receive nearly the same target under either context.

\rparagraph{Problem-specific vs.\ format-induced signal.} The teacher's reasoning format may drive the alignment and not just the content of trajectories. We isolate this effect with the \textit{mismatched} context, which retains the teacher prompt but supplies a solution to a different problem. Such a context produces little relevant signal in vision-language problems (0.04 vs.\ 0.15), and in a different direction (cosine 0.34), which shows that trajectories drive the learning. For text, \textit{mismatched} retains half of the signal (0.09 vs.\ 0.19), in a more similar direction (cosine 0.63): the teacher prompt itself drives much of the signal. This, we believe, is explained by the teacher's ``thinking mode'' reasoning, which the student has not been exposed to \citep{zhao2026self}.\footnote{Since all contexts share the same ``reasoning format'', the raw text cosines are higher between all contexts.} 
\section{Conclusion}
In this work, we study OPSD with \textit{privileged context}, with the teacher instance conditioned on a reasoning trajectory from a stronger expert. We ask \textit{whether that trajectory must be correct}. Our experiments across text-only and vision-language reasoning give a negative answer: \textit{conditioning on correct or incorrect expert trajectories yields similar students}, with gains concentrated on problems the expert solves and the base model does not. We find that the same does not hold for SFT, where directly imitating an incorrect trajectory propagates errors into the student. 
Finally, we probe the teacher's signal and find that correct and incorrect contexts produce nearly the same probability updates on most of the student's tokens. 
Our results uncover an unexpected robustness of OPSD that has a significant practical advantage: it removes the need to (LLM-)judge expert trajectories for correctness.  
%
\clearpage
\section*{Limitations}
First, our incorrect trajectories come from problems the expert solves at least once, so they are likely near-misses. We do not train on problems from the training data that the expert never solves. While our results suggest that self-distillation can use expert trajectories without verifying them (which would remove the need for reference answers), testing this on unfiltered problems and truly unverified problems, as well as problems beyond the expert's reach, is a natural next step.
Second, in both the text-only and the vision-language setting, we use a single model family (Qwen), a single model size of 4B parameters, and the same type of reasoning problems (math). Expanding our work beyond these would test the generality of our findings.
Third, the teacher-signal probe (\S\ref{sec:probe}) measures the signal at initialization, on fixed rollouts of the base model, while during training we sample rollouts on-policy. Tracking the teacher-signal probe as the student trains would show whether the agreement between contexts persists in later stages of the training. Further, this agreement is high but not complete, and the remaining disagreement concentrates on the final tokens of the rollout. We therefore do not claim that the two contexts produce identical teacher targets, only that their shared component dominates the loss. In text, much of the signal is induced by the teacher prompt rather than the trajectory, and most rollouts are truncated before an answer, so the answer has little opportunity to matter by construction.
\bibliography{custom}

\clearpage
\appendix
\section*{Supplementary material}
We organize the appendix as follows:
\begin{itemize}
    \item In \S\ref{app:implementation} we report implementation details: model configurations, training hyperparameters, trajectory pool construction, on-policy rollout setup, and evaluation protocol.
    \item In \S\ref{app:related} we provide an overview of the related work.
    \item In \S\ref{app:llm-judge} we report the LLM as judge details.
    \item In \S\ref{app:ai} we report how we use AI tools in our development.
\end{itemize}
\section{Implementation Details}\label{app:implementation}

\paragraph{Models.}
For text-only experiments, we use Qwen3-4B \citep{qwen3technicalreport} as the student---a hybrid thinking model that can generate with or without a thinking trace depending on the chat template as per \citet{zhao2026self}. For vision-language, we use Qwen3-VL-4B-Instruct with a frozen vision encoder. In both cases, we train only LoRA \citep{hu2022lora} adapters on the language model (rank $r{=}64$, $\alpha{=}128$, dropout $0.1$, applied to \texttt{q,k,v,o,gate,up,down\_proj}). The expert providing the privileged context is Qwen3.6-35B-A3B.

\paragraph{Training.}
All runs train for a single epoch on the filtered training set.
For text-only experiments, we use a learning rate of $5{\times}10^{-6}$, no weight decay, no warmup, gradient norm clipping at $0.1$, and a batch size of $8$ per GPU. We conduct all experiments on a node with 8 $\times$ A100 GPUs. The KL temperature is $1.1$, and we apply OPSD's pointwise divergence clipping at $0.05$ \citep{zhao2026self}. Importantly, we observe extreme sensitivity of the student training to this hyperparameter, as in other works \cite{kaur2026rethinking, shen2026purified, peng2026diagnosing}. The student generates without a thinking trace (\texttt{enable\_thinking=False}), while the teacher prompt uses the thinking template. On-policy rollouts are capped at $1{,}024$ new tokens and the maximum context length is $8{,}192$ tokens. The LoRA adapter is pushed to the vLLM server every $2$ optimizer steps.
For multimodal, we use a learning rate of $5{\times}10^{-5}$  with $3\%$ cosine warmup, gradient norm clipping at $1.0$, and a batch size of 8 samples per GPU. Since we use an instruct-only student, neither a thinking template nor divergence clipping is used ($\tau{=}1.0$). We train with a maximum context length of $8{,}192$ tokens and remove from the training set any samples that exceed that budget. To ensure consistent comparisons, we use the same hyperparameters for the SFT baselines as their self-distillation counterparts, changing only the loss function.
In both settings, we use the AdamW optimizer with a cosine learning-rate schedule, gradient checkpointing, and FSDP2 with BF16 mixed precision.

\paragraph{Trajectory pool.}
For each training problem, we sample $k{=}3$ trajectories from the expert (Qwen3.6-35B-A3B) and $k{=}1$ from the base student (Qwen3-4B or Qwen3-VL-4B-Instruct, before training). Trajectories are sampled with temperature $0.7$, top-$p$ $0.8$, top-$k$ $20$, and presence penalty $1.5$ (non-thinking mode), as per reported best practices \cite{qwen3technicalreport}. We use a maximum of $8{,}192$ new tokens. We label each trajectory as correct or incorrect by comparing its extracted final answer against the reference answer using an LLM judge (Qwen3.6-35B-A3B). To remove the sample-size confounder (\S\ref{sec:method}), we keep only problems for which at least one correct and one incorrect expert trajectory exist, and every condition trains on the same $N$ problems in a single pass.


\paragraph{Evaluation.}
We evaluate text models on AIME24, AIME25, and HMMT Feb 2025, using avg@$12$ (i.e., we sample $12$ responses per problem and report the mean per-problem accuracy). We evaluate in thinking mode with up to $38{,}912$ new tokens (temperature $1.0$, top-$p$ $0.95$, no presence penalty), following \citet{zhao2026self}.

Multimodal models are evaluated on MathVerse \citep{zhang2024mathverse}, MathVision \citep{wang2024measuring}, MathVista \citep{lu2024mathvista}, WeMath \citep{qiao2024we}, and DynaMath \citep{zou2024dynamathdynamicvisualbenchmark}, using avg@$3$ (temperature $0.7$, top-$p$ $0.8$, non-thinking mode, up to $8{,}192$ new tokens). All evaluations use the same LLM judge (Qwen3.6-35B-A3B) for strict answer comparison.

\section{Related Work}\label{app:related}

\paragraph{On-Policy Self-Distillation.}
On-Policy Self-Distillation (OPSD) trains a model on its own rollouts with token-level targets from the same model conditioned on privileged information that is unavailable at inference \cite{zhao2026self, shenfeld2026self, penaloza2026privileged}. This privileged information takes on many forms: reference solutions \cite{zhao2026self}, demonstrations \cite{shenfeld2026self}, action-only trajectories where the internal reasoning is hidden \cite{penaloza2026privileged}, a model's own response together with its binary reward \cite{he2026self}, or several such views combined \cite{nguyen2026avsd}.
In all of these works, the privileged information is either verified against a ground truth or is the ground truth itself. On the other hand, we condition the teacher on trajectories that may be wrong, and measure how much these trajectories, concluding with a final answer, matter for downstream task performance.

\paragraph{Learning from expert reasoning.}
Closest to our setting are works that use a reasoning trajectory from a stronger source (e.g., larger expert model) as the privileged context.
\citet{mendes2026making} train the model to internalize expert human solutions with self-distillation (which are out-of-distribution for the model), by first rewriting them into in-distribution reasoning traces and then training with self-distillation.
\citet{shenfeld2026self} show that conditioning the teacher on a demonstration (which can be a reference solution), rather than fine-tuning on it, forgets less than SFT. Our SFT comparison follows the same logic, but varies the correctness of the demonstrations (e.g., the teacher context).
In contrast to these works, which all assume correct expert content, we show that an incorrect expert trajectory is nearly as useful as a correct one as privileged context, but not as an SFT target.

\paragraph{What the teacher signal contains.}
Several works analyze the per-token signal of OPSD and find that it is not uniformly useful. Several works find that privileged self-distillation can degrade thinking models on long reasoning traces \cite{kaur2026rethinking, peng2026diagnosing}. To fix this issue, \citet{shen2026purified} decompose the teacher's signal into a reference-induced component and a question-conditioned component, and remove the former. Others \cite{xu2026tip} select which tokens to supervise by student entropy and teacher--student divergence, by whether the teacher's correction is reachable from the student's own top candidates \cite{wang2026not}, or by student confidence \cite{zhang2026robust}; \citet{zhao2026self} instead clip the per-token divergence, which we find to be necessary for OPSD to work in the specific text-only setting they propose. The way we probe the teacher is complementary (\S\ref{sec:probe}): we compare the signals induced by different contexts for the same rollout, and find that correct and incorrect trajectories induce nearly the same update.

\paragraph{Self-distillation for vision-language models.}
In vision-language models, the privileged signal is often visual rather than textual. A teacher can see a zoomed crop of the relevant region \cite{yuan2026vision} or visual cues that the student can recover from its own input \cite{tian2026vicur}, or receive structured hints built from spatial attention and intermediate reasoning steps \cite{xiang2026teaching}. Other works create the teacher--student asymmetry by exposing the student to less information than the teacher (thus training the student to be invariant to such missing information): masking the student's most-attended reasoning prefixes \cite{yu2026hide}, augmenting the student's view of the image \cite{li2026self}, or contrasting the teacher's distributions with and without the image content \cite{liang2026visual}. Several on-policy distillation methods target visual grounding directly \cite{liu2026visual, yoon2026decomposed}, including without answer labels \cite{sun2026v}. In contrast to these works, our goal is to measure the impact of the correctness of the privileged information on OPSD, in both text and vision-language settings.

\paragraph{Training on unverified data.}
Training on unverified data can improve reasoning on its own: SFT on a model's own samples improves math reasoning without rewards or verifiers \cite{li2025model}, and fine-tuning on unverified samples improves code generation \cite{zhang2026embarrassingly}. Further, \citet{lee2026self} filter self-generated solutions with prompt-based self-verification checks before training on them. Instead, we ask whether the trajectory given to the teacher needs to be verified at all. This is consistent with the modest gains we observe when training on the base model's own trajectories, and is why we compare expert and self contexts under identical problems and budgets.

\section{LLM-as-judge}\label{app:llm-judge}
All LLM-as-judge evaluations---both for building the trajectory pool and for scoring evaluation predictions---use a non-thinking Qwen3.6-35B-A3B with temperature 0. The judge responds with a single token (\texttt{YES} or \texttt{NO}); failures default to \texttt{NO} (an ungradable response counts as incorrect).

\paragraph{Trajectory annotation (training pool).}
To label each sampled trajectory as correct or incorrect (\S\ref{sec:method}), we compare the trajectory's extracted final answer against the reference answer. The judge receives only the predicted and ground-truth answer (no question text) and decides semantic equivalence under the following rules:
\begin{itemize}
    \item \textbf{Multiple-choice:} If the ground-truth is a letter A--E (optionally prefixed by ``Answer:'' or ``The answer is''), the prediction matches when it begins with the same letter, regardless of any trailing value or expression.
    \item \textbf{Numeric:} Values that are equal after rounding or unit-stripping match (e.g., ``584.2'' vs.\ ``584.2 square inches'', ``0.5'' vs.\ ``1/2'').
    \item \textbf{Latex:} A ground-truth wrapped in \texttt{\textbackslash boxed\{...\}} is compared by its inner content only.
    \item \textbf{Synonyms:} Trivial wording differences and paraphrases that refer to the same entity match (e.g., ``a dog'' vs.\ ``dog'').
\end{itemize}
We also use a mode that passes the last 1,500 characters of the reasoning trace instead of the extracted answer, letting the judge locate the final answer itself. This is robust to extraction failures (e.g., separator lines or LaTeX artifacts captured as the ``answer'' by the last-line heuristic).

\paragraph{Evaluation scoring.}
During inference, the judge receives the \textit{full question} (with answer choices), the \textit{entire model response}, and the ground-truth answer. Then, it locates the model's final answer within the response (e.g., in a \texttt{\textbackslash boxed\{...\}}, on a ``Final Answer:'' line) and applies the same equivalence rules as above. Supplying the question lets the judge map a bare numeric answer (e.g., ``49'') onto its lettered choice (e.g., ``A: 49'') when the ground-truth is a letter (e.g., in the case of multiple-choice answers), resolving the main failure mode of the question-free variant. We truncate long responses to 12,000 characters before judging (the final answer is at the end).

\section{Use of AI assistants}\label{app:ai}
We used AI assistants for coding (implementing training, evaluation, and analysis scripts) and for writing (e.g., drafting and editing text). We, the authors, designed all experiments, ran them, and checked all reported numbers against the experiment outputs. We also reviewed and edited all AI-assisted code and text, verified every citation against the cited paper, and take full responsibility for the content of the paper.

\end{document}